\documentclass[letterpaper]{article} 
\usepackage{aaai2026}  
\usepackage{times}  
\usepackage{helvet}  
\usepackage{amsfonts}
\usepackage{amsmath}
\usepackage{courier}  
\usepackage[hyphens]{url}  
\usepackage{algorithm}

\usepackage{graphicx} 
\usepackage{booktabs}
\usepackage{tabularx}  
\usepackage{multirow}

\usepackage{natbib}  
\usepackage{caption} 
\usepackage{booktabs} 
\usepackage{algorithm}
\usepackage{algorithmic}

\usepackage{newfloat}
\usepackage{listings}
\DeclareCaptionStyle{ruled}{labelfont=normalfont,labelsep=colon,strut=off} 
\floatstyle{ruled}
\newfloat{listing}{tb}{lst}{}
\floatname{listing}{Listing}
\title{Where and What Matters: Sensitivity-Aware Task Vectors for \\Many-Shot Multimodal In-Context Learning}
\author{
    Ziyu Ma\textsuperscript{\rm 1}\equalcontrib, Chenhui Gou\textsuperscript{\rm 2}\equalcontrib, Yiming Hu\textsuperscript{\rm 1}\thanks{Project leads and corresponding authors.}, Yong Wang\textsuperscript{\rm 1}\footnotemark[2], Xiangxiang Chu\textsuperscript{\rm 1}, \\Bohan Zhuang\textsuperscript{\rm 3}, Jianfei Cai\textsuperscript{\rm 2}
}
\affiliations{
    \textsuperscript{\rm 1}AMAP, Alibaba Group \textsuperscript{\rm 2}Monash University
    \textsuperscript{\rm 3}Zhejiang University
}

\usepackage{bibentry}
\usepackage{color}
\begin{document}

\maketitle

\begin{abstract}
Large Multimodal Models (LMMs) have shown promising in-context learning (ICL) capabilities, but scaling to many-shot settings remains difficult due to limited context length and high inference cost. To address these challenges, task-vector-based methods have been explored by inserting compact representations of many-shot in-context demonstrations into model activations. However, existing task-vector-based methods either overlook the importance of where to insert task vectors or struggle to determine suitable values for each location. To this end, we propose a novel Sensitivity-aware Task Vector insertion framework (STV) to figure out \textit{where and what} to insert. Our key insight is that activation deltas across query-context pairs exhibit consistent structural patterns, providing a reliable cue for insertion. Based on the identified sensitive-aware locations, we construct a pre-clustered activation bank for each location by clustering the activation values, and then apply reinforcement learning to choose the most suitable one to insert. We evaluate STV across a range of multimodal models (e.g., Qwen-VL, Idefics-2) and tasks (e.g., VizWiz, OK-VQA), demonstrating its effectiveness and showing consistent improvements over previous task-vector-based methods with strong generalization. Our code will be available at \url{https://github.com/AMAP-ML/STV}.

\end{abstract}

\section{Introduction}
Large Multimodal Models (LMMs), such as GPT-4V \cite{openai2023gpt}, LLaVA \cite{li2024llava,liu2024improved}, and Qwen-VL \cite{wang2024qwen2, bai2025qwen2}, have demonstrated strong in-context learning (ICL) capabilities for vision-language tasks, particularly in few-shot scenarios \cite{bai2023qwen, laurenccon2024matters}. Recent findings from commercial LMMs including Gemini \cite{comanici2025gemini} and GPT-4o \cite{hurst2024gpt}, further highlight the potential of many-shot ICL, where increasing the number of in-context examples markedly enhances performance on multimodal tasks \cite{agarwal2024many, jiang2024many}. However, leveraging many-shot ICL in open-source LMMs remains challenging due to two key limitations: (1) limited context length (e.g., Qwen-VL \cite{bai2023qwen} supports a maximum of 8192 tokens, but encoding a single image already consumes up to 256 tokens), and (2) substantial inference overhead, as incorporating more examples increases both memory usage and inference latency \cite{ma2025drvideo}.

\begin{figure}[t]
\centering
\includegraphics[width=1\columnwidth]{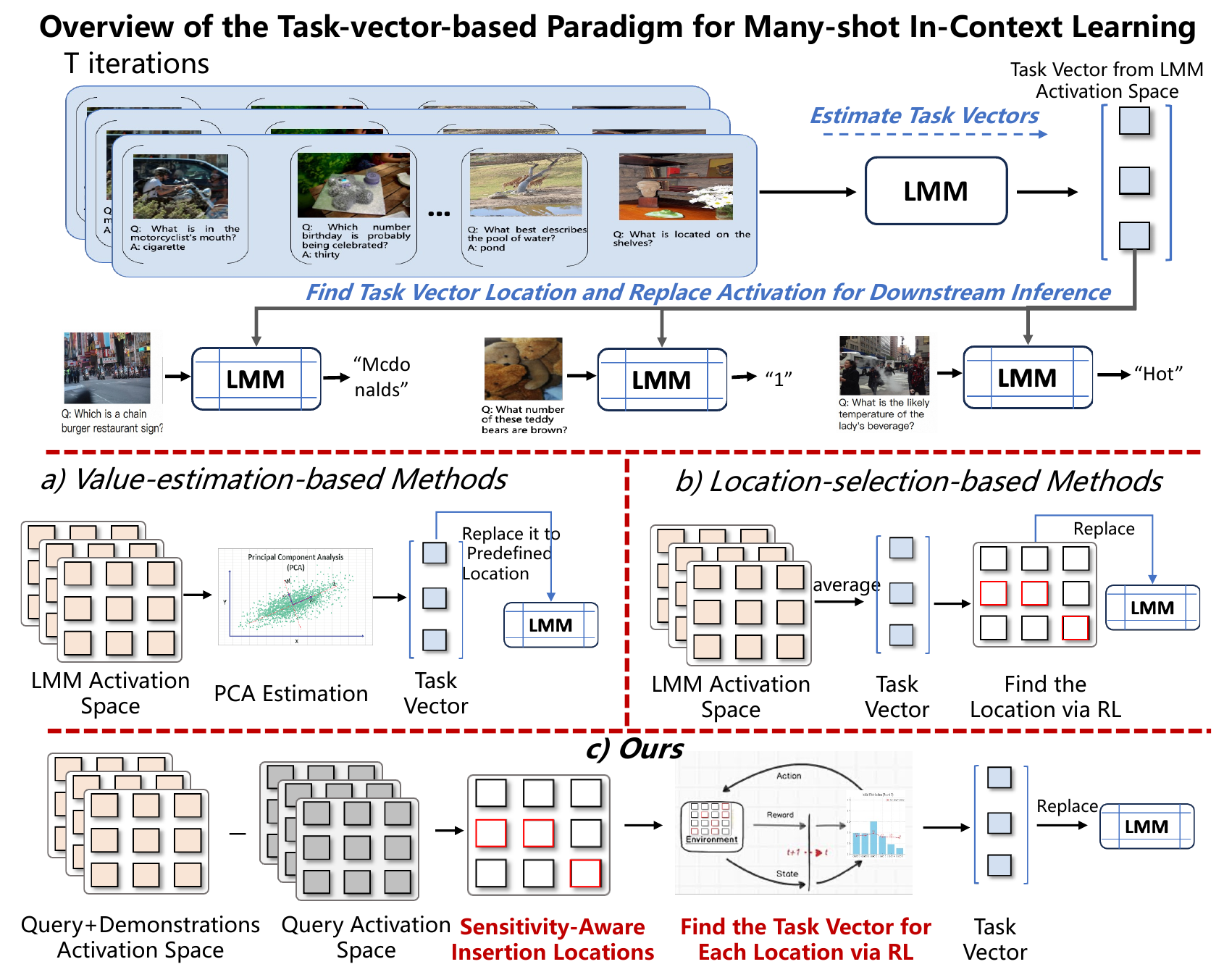} 
\caption{Comparison of previous task-vector-based methods and our sensitivity-aware method that systematically determines both insertion locations and task vector values.}
\label{fig1-1}
\end{figure}



\begin{figure*}[t]
\centering
\includegraphics[width=2.0\columnwidth]{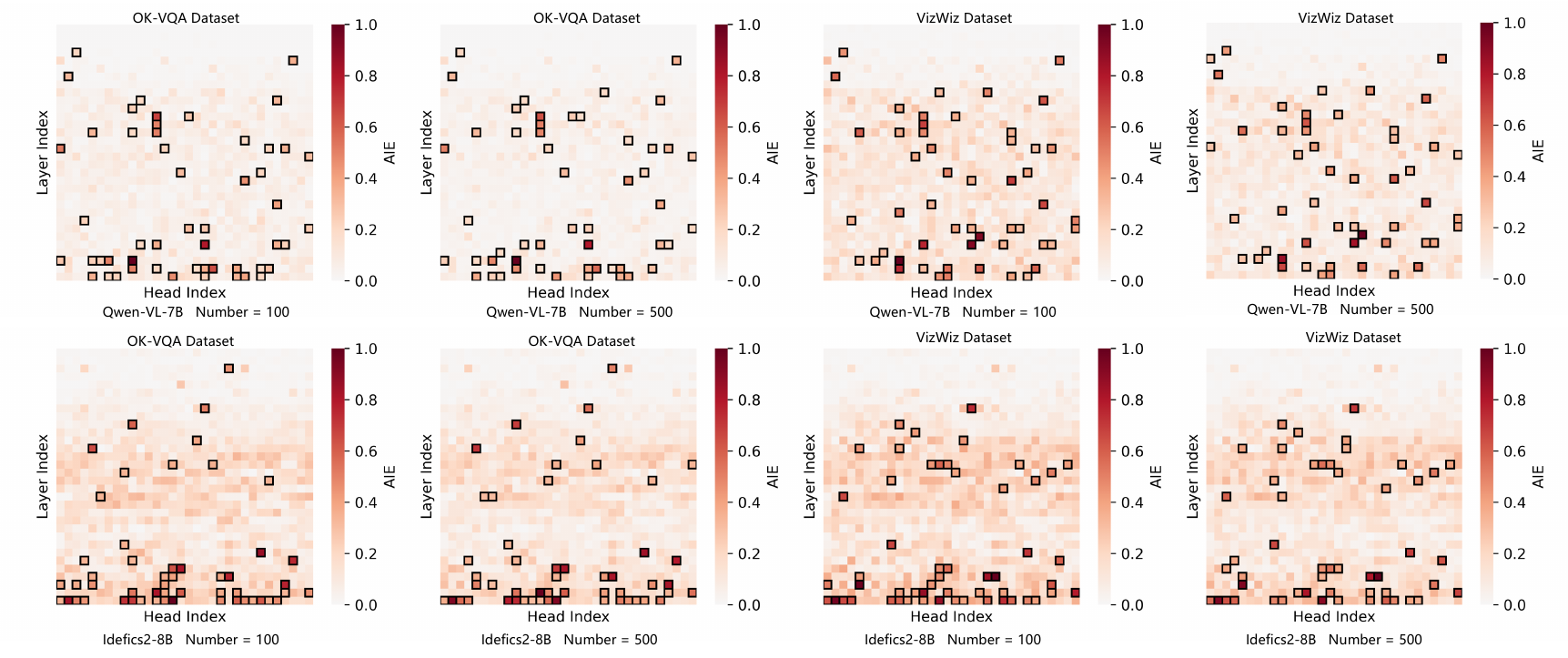} 
\caption{Attention head sensitivity across datasets (VizWiz and OK-VQA), models (Qwen-VL-7B and idefics2-8B), and sample sizes (100 vs. 500). Context-sensitive locations (black boxes) consistently emerge within tasks, validating the stability and structural patterns of activation delta, which are computed by contrasting query–context activations with query-only activations.}
\label{fig2-1}
\end{figure*}


The paradigm based on task vectors has recently emerged as a promising solution, attracting increasing attention. Unlike conventional methods that concatenate all in-context demonstrations into the input sequence, task-vector techniques \cite{hendel2023context,liu2023context, todd2023function, huang2024multimodal} compress many-shot demonstrations into compact, semantically rich representations (i.e., task vectors). They are then integrated into the model's attention heads to address downstream tasks. By reframing the problem, this paradigm shifts focus from managing long input sequences to addressing two fundamental challenges: \textit{(1) how to construct semantically meaningful task vectors, and (2) where within the model architecture to insert them for optimal effectiveness.}

Previous studies \cite{hendel2023context, liu2023context, todd2023function} have introduced the \textit{value-estimation-based} methods (see Fig. \ref{fig1-1}(a)), which primarily focus on deriving task vectors from demonstrations and inserting them into predefined locations within the model. A representative approach \cite{liu2023context} employs Principal Component Analysis (PCA) \cite{abdi2010principal} on hidden states from multiple demonstrations and a dummy query, producing compact vectors. Such vectors are then used to replace the representation of the real query at fixed attention layers during inference. While this strategy mitigates the challenge of processing long input sequences and demonstrates efficacy for simple tasks, it  struggles to generalize to complex, multimodal tasks \cite{NEURIPS2024_12d3e63b}. Furthermore, it does not account for the critical influence of insertion locations on task performance. In contrast, recent approaches \cite{huang2024multimodal} shift focus to optimizing insertion locations while using fixed task vectors, a framework known as the \textit{location-selection-based} method (see Fig. \ref{fig1-1}(b)). They typically compute the mean activation across demonstrations to create a task vector, which is then used by a policy network to identify the insertion point for maximizing performance. However, these methods often harm the representation richness of task information and lead to the loss of task-specific details due to averaging multiple in-context demonstrations into a single fixed vector. Moreover, insertion typically relies on sampling-based policies, and we observe that repeated runs yield inconsistent
locations, and some locations have little or no impact
on model predictions, indicating suboptimal insertion.

The inherent limitations of existing methods raise a fundamental question: \textit{Is it possible to determine where and what to insert in a more systematic and reliable manner?}


\begin{figure*}[t]
\centering
\includegraphics[width=0.9\textwidth]{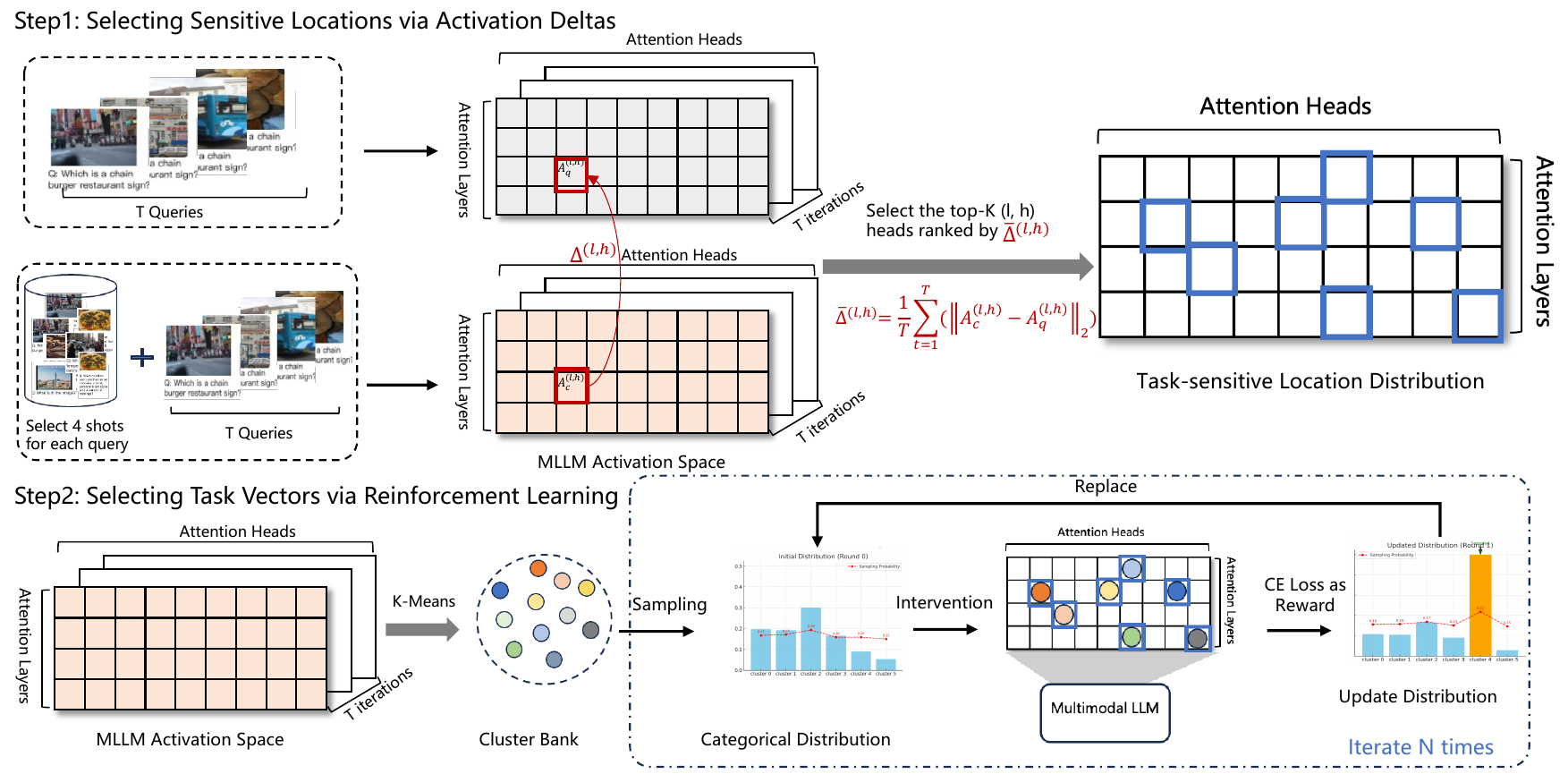} 
\caption{Overview of the STV Framework. It consists of two stages: (1) Sensitivity-aware location identification, where we compare query–context activations with query-only activations to compute activation deltas, and determine attention heads that consistently respond to contextual information. (2) Task vector selection, where candidate task vectors are drawn from a pre-computed activation bank, and reinforcement learning is used to choose the most suitable one at the identified locations.}
\label{framework}
\end{figure*}

To address this issue, we propose a novel framework, Sensitivity-aware Task Vector insertion (STV), which determines \textit{where} to insert and \textit{what} to insert when incorporating task vectors. This framework is inspired by our observation that the effective insertion locations exhibit consistent, task-dependent patterns (see Section 2.1 for details). Based on this, STV consists of two stages: (1) identifying locations that are sensitive to contextual information, and (2) selecting optimal task vectors for insertion. In the first stage, we quantify contextual sensitivity by computing activation deltas between queries with and without demonstrations. Specifically, averaging these deltas across multiple samples highlights stable patterns, and the top-$k$ locations with the highest sensitivity scores are chosen as insertion points. In the second stage, we first construct a pre-clustered activation bank for each location by clustering activation values derived from multiple query-context forward passes. For each identified sensitive location, a candidate task vector is then sampled from this activation bank by a learnable discrete distribution. And this distribution is iteratively updated through the REINFORCE algorithm \cite{williams1992simple}, enabling the model to progressively identify the most effective vector for each location.

We systematically evaluate our method across multiple LMM families, including Qwen-VL \cite{bai2023qwen}, and Idefics-2 \cite{laurenccon2024building}, as well as a diverse set of multimodal benchmarks. The results demonstrate that our approach consistently outperforms the previous state-of-the-art task-vector-based method, MTV \cite{huang2024multimodal}, while reducing location-searching time by over 98\%. Besides, compared to traditional fine-tuning approaches, such as LoRA \cite{hu2022lora}, our method achieves superior accuracy with significantly lower computational resource requirements. Notably, it operates in an end-to-end manner on a single GPU with less than 20GB of memory, outperforming both zero-shot and few-shot ICL baselines \cite{brown2020language} without requiring any model finetuning.

Our contributions can be summarized as follows:

\begin{itemize}
\item We observe that activation deltas exhibit consistent, task-dependent patterns within a given model, revealing structurally stable sensitivity to contextual information. Building on this insight, we propose a novel and efficient sensitivity-aware strategy to identify insertion locations.
\item In contrast to previous methods that rely on PCA estimation or mean activations, we propose a policy-driven strategy that progressively selects the most effective task vector for each insertion location through RL.
\item Empirical results across five vision-language tasks and two LMMs, consistently show that our method achieves superior performance, outperforming existing task-vector-based methods.
\end{itemize}

\section{Method}
We propose a Sensitivity-aware Task Vector insertion framework (STV) to enable efficient many-shot in-context learning in LMMs without extending context length. We begin by describing the background on multimodal ICL and task vectors, followed by our observations (Section 2.1). Then our two-step method is introduced: (1) identifying where to insert task vectors by measuring attention head sensitivity to in-context examples, and (2) selecting what to insert by sampling from a pre-clustered activation bank via reinforcement learning. The overall framework is illustrated in Fig.~\ref{framework}.


\subsection{Preliminary Study}

In the multimodal in-context learning (ICL) setting, a large multimodal model (LMM) learns to perform a new task based on a few interleaved image-text examples without parameter updates. The input typically takes the form:
\begin{equation}
I_{\text{few}} = \{(x_1, y_1), (x_2, y_2), \ldots, (x_n, y_n), q\},
\label{eq:ifew}
\end{equation}
where $(x_i, y_i)$ denotes the multimodal input-output pairs and $q$ is the test query. Instead of concatenating all examples into the input, task-vector-based methods aim to represent these demonstrations more compactly. Specifically, the task vector is defined as a tuple $(\mu_j, \lambda_j)$, where $\mu_j$ is a set of activation vectors capturing task-level semantics. $\lambda_j$ is a set of attention head locations (indexed by layer and head) where such vectors are to be inserted during inference.

Given a model $F$, let $\lambda = \{(l, h)\}$ denotes the full set of attention heads, where $l$ is the layer index and $h$ is the head index. For each head $(l, h)$, the output before linear projection is denoted as $A^{(l,h)} \in \mathbb{R}^d$, which we refer to as the \emph{activation}. Previous works directly insert $\mu_j$ into $\lambda_j$ to modulate model behavior during inference.

However, we observe that the effective locations $\lambda_j$ for inserting task vectors vary significantly across \textit{tasks} and \textit{models}. By computing the \textit{activation delta} (i.e., the difference between a query’s activation with and without demonstrations), we find that locations with large deltas are consistently distributed within the same model (e.g., Qwen-VL \cite{bai2023qwen}) and task (e.g., OK-VQA \cite{marino2019ok}), as shown in Fig.~\ref{fig2-1}, indicating high sensitivity to contextual information and revealing structural patterns in how in-context is processed. On the other hand, these sensitive locations vary notably across models (e.g., Qwen-VL vs. idefics-2 \cite{laurenccon2024building}) and tasks (e.g., VizWiz \cite{gurari2018vizwiz} vs. OK-VQA). The above observations form the core motivation of our method, which leverages activation deltas to identify context-sensitive locations for efficient task vector insertion.


\subsection{Activation Delta for Location Selection}

Given a query $q$, its activation at each attention head $(l, h)$ is denoted as $A_q^{(l,h)} \in \mathbb{R}^{1 \times d}$. A context-augmented input is constructed by concatenating $S$ in-context examples with the query, as defined in Eq.~\eqref{eq:ifew}, i.e., $c = I_{\text{few}}$, which produces activations $A_c^{(l,h)}$.

To estimate the sensitivity of each location, we measure the L2 activation difference between the query with and without context. Specifically, it is computed as:
\begin{equation}
\bar{\Delta}^{(l,h)} = \frac{1}{T} \sum_{t=1}^{T} \left\| A_{c_t}^{(l,h)} - A_{q_t}^{(l,h)} \right\|_2,
\label{eq:avg_delta}
\end{equation}
where \((A_{c_t}^{(l,h)}, A_{q_t}^{(l,h)})\) are activations from the $t$-th sampled query-context pair. Here, averaging over $T$ iterations ensures stable estimation \cite{he2023sensitivity}.

The resulting matrix $\bar{\Delta} \in \mathbb{R}^{L \times H}$ captures the context sensitivity distribution across all attention heads. Then the top-$K$ heads with the highest values in $\bar{\Delta}$ is selected as the sensitive locations:
\begin{equation}
\Lambda = \mathrm{TopK}(\bar{\Delta}, K)
\label{eq:topk},
\end{equation}
which serves as the basis for the second stage, where appropriate task vectors are determined and inserted.

\subsection{Selecting Task Vectors via RL}

Given the set of sensitivity-informed insertion locations $\mathcal{L} = \{(l_k, h_k)\}_{k=1}^K$, the next step is to determine appropriate activation values to insert at each location. A pre-clustered activation bank is constructed by collecting context-enhanced activations across multiple examples and applying offline clustering \cite{krishna1999genetic}. 

For each location $(l_k, h_k)$, $M$ cluster centers are computed, forming a discrete candidate set:
\begin{equation}
\texttt{ClusterBank}[(l_k, h_k)] = \{ \mathbf{v}_1^{(k)}, \ldots, \mathbf{v}_M^{(k)} \}, \quad \mathbf{v}_i^{(k)} \in \mathbb{R}^d.
\label{eq:clusterbank}
\end{equation}

Selecting task vectors from this bank is framed as a discrete optimization problem. For each location, a categorical distribution is defined over the $M$ candidates, parameterized by learnable logits $\boldsymbol{\alpha}^{(k)} \in \mathbb{R}^M$:
\begin{equation}
\mathbf{p}^{(k)} = \mathrm{softmax}(\boldsymbol{\alpha}^{(k)}).
\label{eq:softmax}
\end{equation}

During training, cluster indices $\{i_k\}$ are sampled from the distributions $\mathbf{p}^{(k)}$, forming a full vector set $\mathcal{V}_i = \{ \mathbf{v}_{i_k}^{(k)} \}_{k=1}^K$. Then they are inserted into the model activations at the corresponding locations $\Lambda$ during inference. The resulting task loss $\mathcal{L}_{\mathrm{task}}$ on sample $(x_i, y_i)$ is used to compute a reward:
\begin{equation}
r_i = -\mathcal{L}_{\mathrm{task}}(F(x_i; \Lambda, \mathcal{V}_i), y_i).
\label{eq:reward}
\end{equation}

Policy learning is performed using the REINFORCE algorithm \cite{williams1992simple}. The policy loss is defined as:
\begin{equation}
\mathcal{L}_{\mathrm{policy}} = - \sum_{i=1}^{N} \sum_{k=1}^K \log p^{(k)}_{i_k} \cdot \frac{r_i - \bar{r}}{\sigma_r + \epsilon},
\label{eq:policy}
\end{equation}
where $\bar{r}$ and $\sigma_r$ are moving averages of reward mean and standard deviation for variance reduction.

The optimization is conducted via gradient descent on $\{\boldsymbol{\alpha}^{(k)}\}$ to encourage sampling high-reward clusters. After convergence, the final task vectors are determined by choosing the highest-probability cluster at each location:
\begin{equation}
\hat{\mathbf{v}}^{(k)} = \mathbf{v}_{i_k^*}^{(k)}, \quad i_k^* = \arg\max_i \alpha_i^{(k)}.
\label{eq:final_vector}
\end{equation}
In this way, this RL-based vector selection enables STV to flexibly adapt to downstream tasks while avoiding exhaustive enumeration over the candidate space.

\subsection{Inference with Task Vector Modification}

Given the selected insertion locations $\mathcal{L} = \{(l_k, h_k)\}_{k=1}^K$ and task vectors $\hat{\mathcal{V}} = \{\hat{\mathbf{v}}^{(k)}\}_{k=1}^K$, inference is conducted by modifying intermediate activations at the designated locations during the forward pass.

For a test input $x_{\text{test}}$, the original model activations at each $(l_k, h_k)$ are replaced with the corresponding task vector $\hat{\mathbf{v}}^{(k)}$. Let $F^\dagger$ denote the modified model under task vector intervention, the prediction is computed as:
\begin{equation}
\hat{y}_{\text{test}} = F^\dagger(x_{\text{test}}; \mathcal{L}, \hat{\mathcal{V}}).
\end{equation}
This enables many-shot conditioning at the activation level without increasing the input length or introducing additional parameters. Only a single forward pass is required per test instance, allowing efficient and scalable inference under various input settings. 

\section{Experiments}

\begin{table*}[t]
\centering
\renewcommand{\arraystretch}{1.12}
\setlength{\tabcolsep}{6pt}
\begin{tabular}{l|l|*{5}{m{1.5cm}}|c|c}
\toprule
\textbf{Category} & \textbf{Method} & \textbf{VizWiz} & \textbf{OK-VQA} & \textbf{DTD} & \textbf{Flowers} & \textbf{CUB} & \textbf{Avg.} & \textbf{Avg. Gain} \\
\midrule
\multicolumn{9}{c}{\textbf{\textit{Qwen-VL-7B Models}}} \\
\midrule
\multirow{2}{*}{Standard ICL} 
& zero-shot ICL & 35.21 & 57.76 & 55.07 & 55.24 & 56.50 & 51.96 & -- \\
& 4-shot ICL   & 42.00 & 54.62 & 55.50 & 54.67 & 56.16 & 52.59 & $\uparrow$0.63 \\
\midrule
\multirow{6}{*}{\parbox{2.2cm}{\centering Task Vector\\Based Methods}}
& TV            & 36.71 & 40.06 & 50.93 & 49.78 & 52.89 & 46.07 & $\downarrow$5.89 \\
& FV            & 40.51 & 52.78 & 53.23 & 52.95 & 53.56 & 50.61 & $\downarrow$1.35 \\
& ICV           & 37.24 & 55.89 & 49.56 & 50.04 & 51.64 & 48.87 & $\downarrow$3.09 \\
& I2CL          & 39.39 & 56.67 & 67.71 & 59.63 & 62.53 & 57.99 & $\uparrow$6.03 \\
& MTV           & \underline{45.60} & \underline{60.51} & \underline{76.50} & \underline{78.10} & \underline{80.00} & \underline{68.94} & $\uparrow$16.98 \\
& \textbf{STV (Ours)} & \textbf{58.30} & \textbf{61.94} & \textbf{80.45} & \textbf{81.51} & \textbf{82.33} & \textbf{72.11} & $\uparrow$\textbf{20.15} \\
\midrule
\multicolumn{9}{c}{\textbf{\textit{Idefics2-8B Models}}} \\
\midrule
\multirow{2}{*}{Standard ICL} 
& zero-shot ICL & 31.30 & 52.40 & 88.73 & 82.80 & 88.70 & 68.79 & -- \\
& 4-shot ICL   & 40.80 & 51.50 & 89.13 & \underline{84.32} & 87.26 & 70.60 & $\uparrow$1.81 \\
\midrule
\multirow{6}{*}{\parbox{2.2cm}{\centering Task Vector\\Based Methods}}
& TV            & 28.82 & 41.67 & 62.33 & 50.97 & 62.16 & 49.59 & $\downarrow$19.20 \\
& FV            & 26.79 & 44.51 & 56.53 & 54.36 & 67.73 & 49.18 & $\downarrow$19.61 \\
& ICV           & 25.59 & 38.62 & 52.37 & 48.64 & 60.46 & 45.94 & $\downarrow$22.85 \\
& I2CL          & 29.03 & 47.13 & 73.63 & 62.17 & 74.50 & 57.29 & $\downarrow$11.50 \\
& MTV           & \underline{52.50} & \underline{53.00} & \underline{89.14} & 83.80 & \underline{89.80} & \underline{73.65} & $\uparrow$4.86 \\
& \textbf{STV (Ours)} & \textbf{60.61} & \textbf{54.14} & \textbf{92.47} & \textbf{86.73} & \textbf{90.23} & \textbf{76.04} & $\uparrow$\textbf{7.25} \\
\bottomrule
\end{tabular}
\caption{Performance comparison across five datasets. Best results are in \textbf{bold}, second best are underlined. The \textbf{Average Gain} column indicates improvement relative to zero-shot ICL.  The first part reports results of standard ICL under zero-shot and few-shot settings. The second part presents task-vector-based methods including TV, FV, ICV, I2CL, and MTV. The final row shows our STV, which consistently achieves the best performance and the largest average gains across models and datasets.}
\label{aaai_tab:stv_results}
\end{table*}

\begin{figure*}[t]
\centering
\includegraphics[width=0.9\textwidth]{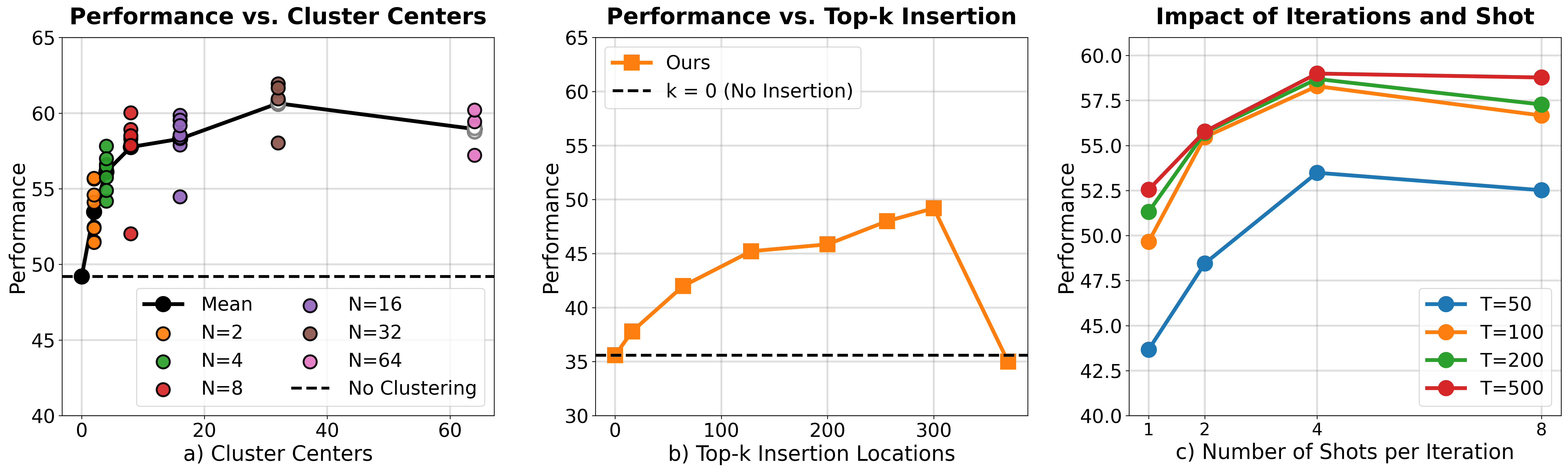} 
\caption{a) Effect of Cluster Granularity on Task Vector Selection using Qwen-VL on VizWiz Dataset. b) Impact of Top-k Location Selection on Model Performance using Qwen-VL on VizWiz Dataset. c) Impact of Iterations $T$ and the Number Shot per Iteration on Model Performance using Qwen-VL on VizWiz Dataset.}
\label{fig4-1}
\end{figure*}


\subsection{Evaluation Setup}

We evaluate our method on five vision-language benchmarks: VizWiz \cite{gurari2018vizwiz}, OK-VQA \cite{marino2019ok}, DTD \cite{cimpoi14describing}, Flowers \cite{nilsback2008automated}, and CUB \cite{wah2011caltech}, covering diverse reasoning types and image domains, including real-world user-captured photos (VizWiz), knowledge-intensive visual questions (OK-VQA), and fine-grained classification tasks across textures, flowers, and bird species (DTD, Flowers, CUB). More details are reported in the Appendix.


\begin{table}[ht]
\centering
\begin{tabular}{l|cc|c}
\toprule
Methods & MTV & Ours & Change \\
\midrule
Location Search Time (s) & 6000 & 88 & $\downarrow$ 98.53\% \\
GPU Memory (GB) & 19.8 & 19.8 & - \\
Inference Time (s) & 0.49 & 0.49 & - \\
Performance & 45.6 & \textbf{58.3} & $\uparrow$ 12.7 \\
\bottomrule
\end{tabular}
\caption{Comparison between MTV and our method on task vector location search efficiency, resource consumption, and performance on VizWiz dataset. }

\label{aaai_tab:comparison}
\end{table}


To assess the cross-model generalization of our method, we test on two interleaved Large Multimodal Models:  
\begin{itemize}
    \item Qwen-VL-7B \cite{bai2023qwen}: 
    a bilingual vision-language model with a context window of 8192 tokens, 
    requiring 256 tokens per image.  
    
    \item Idefics2-8B \cite{laurenccon2024building}: 
    a strong open-source model optimized for high-resolution image understanding and flexible alignment, 
    with a context window of 8192 tokens and 64 tokens per image.
\end{itemize}


All evaluations are performed using consistent in-context prompting without gradient updates. Besides, we compare our method with two categories of baselines:  

\begin{itemize}
    \item Standard In-Context Learning (ICL) methods \cite{brown2020language}, which directly concatenate multimodal exemplars with the query into the model input sequence.
    \item Task-vector-based methods, which compress in-context demonstrations into activation space for efficient adaptation, including TV \cite{hendel2023context}, FV \cite{todd2023function}, ICV \cite{liu2023context}, I2CL \cite{li2024implicit}, and MTV \cite{huang2024multimodal}.  
\end{itemize}

\subsection{Implementation Details}

All experiments are implemented in PyTorch using official model checkpoints for Qwen-VL-7B and Idefic-2-8B. All evaluations are performed on a single NVIDIA H20 GPU unless otherwise specified. For each dataset, we use 100 in-context examples under a 4-shot setting. During inference, we apply zero-shot prompting to isolate the effect of task vector insertion. Unless noted, the maximum generation length is capped at 20 tokens. To identify sensitive locations, we precompute activation deltas and select 64 attention head locations. For each location, we construct a task vector bank by clustering the corresponding activations from many-shot forward passes using $k$-means with 16 clusters. Additional implementation details, including ablation protocols and hyperparameter sensitivity, are reported in the Appendix.

\subsection{Main Results}
Table~\ref{aaai_tab:stv_results} summarizes the performance of our STV compared to a range of strong baselines across 5 benchmarks and 2 representative LMM families. Several key observations emerge:



\noindent\textbf{Performance across VQA Datasets.} 
On VizWiz, STV achieves 58.3\% accuracy with Qwen-VL-7B and 60.6\% with Idefics2-8B, surpassing the strongest baseline MTV by +12.7\% and +8.1\%, respectively. 
On OK-VQA, STV obtains 61.9\% (Qwen-VL-7B) and 54.1\% (Idefics2-8B), exceeding 4-shot ICL by +7.3\% and +2.6\%. 
On fine-grained classification datasets, STV yields 80.5\% on DTD, 81.5\% on Flowers, and 82.3\% on CUB with Qwen-VL-7B, consistently improving over 4-shot ICL by +25.0\%, +26.8\%, and +26.2\%. 
These results indicate STV enhances robustness on noise-rich real-world datasets such as VizWiz, while also excelling on fine-grained recognition benchmarks by leveraging activation sensitivity to capture discriminative cues.

\noindent\textbf{Generalization across Diverse Model Architectures.} 
With Qwen-VL-7B, STV improves the average accuracy from 68.9\% (MTV) to 72.1\% (+3.2\%). 
With Idefics2-8B, STV improves the average accuracy from 73.7\% (MTV) to 76.0\% (+2.4\%). 
Unlike ICV and FV, whose performance fluctuates depending on the underlying architecture, STV maintains leading accuracy across distinct backbone designs (i.e., encoder-centric Qwen-VL-7B and decoder-centric Idefics2-8B). 
This suggests that STV’s  sensitive-aware location and RL-based task vector selection mechanism captures transferable structural cues, ensuring stable adaptation across heterogeneous model families.

\noindent\textbf{Efficiency and Comparative Advantage over Baselines.} 
Compared with MTV, which requires exhaustive location search, STV achieves comparable or better performance while reducing computational overhead to only 1.5\% of MTV's search cost (Table \ref{aaai_tab:comparison}). On Qwen-VL-7B, STV improves the average accuracy by +3.2\%, and on Idefics2-8B by +2.4\%, while also providing more stable gains across datasets. These results indicate that STV improves both performance and efficiency compared to existing task vector methods, offering a balanced and practical solution for multimodal in-context adaptation.

\subsection{Ablation Study}

\noindent\textbf{Effect of Task Vector Construction and Insertion Locations.} 
We study STV’s two key components: \textit{where and what to insert}, on VizWiz with Qwen-VL-7B. 
As shown in Fig.~\ref{fig4-1}(b), performance rises from 35.2\% to 49.2\% as more sensitive locations are used, confirming the effectiveness of our sensitivity-aware selection; yet overly large $k$ ($>300$) sharply degrades accuracy, showing that untargeted intervention disrupts model reasoning. 
Fig.~\ref{fig4-1}(a) further fixes $k=300$ and varies the number of cluster centers $N$: performance improves steadily with larger $N$ and saturates around 32, indicating that clustering yields richer semantic task vectors and mitigates information loss. Together, location selection contributes +14.0\% and task vector construction adds +9.1\%, validating both components as essential to STV.

\noindent\textbf{Scaling In-Context Samples.} 
We further examine the scalability of STV with respect to the number of iterations $T$ and the number of shots per iteration. 
In Fig.~\ref{fig4-1}(c), increasing either $T$ or the number of shots consistently improves performance, demonstrating two key points: (i) STV can effectively process a large number of in-context examples, highlighting the advantage of task vector-based methods over direct concatenation; (ii) larger context provides richer task information, from which STV is able to extract useful cues. 
However, when $T$ or the shot size grows excessively, accuracy begins to drop, suggesting that redundant information introduces noise and interferes with reasoning. 
These trends confirm that STV achieves robust adaptation under larger contexts while benefiting from principled sample scaling.


\noindent\textbf{Comparison with Parameter-Tuning Adaptation Methods.}
We further compare STV with two common adaptation paradigms: LoRA-based parameter-efficient tuning and full supervised fine-tuning (SFT). As shown in Table~\ref{tab:vizwiz_okvqa_comparison}, STV consistently outperforms LoRA on both VizWiz and OK-VQA, while requiring no parameter updates or task-specific optimization. Although SFT achieves strong performance on VizWiz (+26.8), it suffers a severe drop on OK-VQA ($-$33.5), revealing overfitting and poor transferability. This highlights a key limitation of parameter tuning: it yields task-specific gains at the cost of generalization. In contrast, STV achieves robust improvements across domains (+23.1 on VizWiz, +3.3 on OK-VQA), demonstrating the effectiveness of activation-level adaptation without parameter tuning.

\begin{figure}[t]
\centering
\includegraphics[width=0.9\columnwidth]{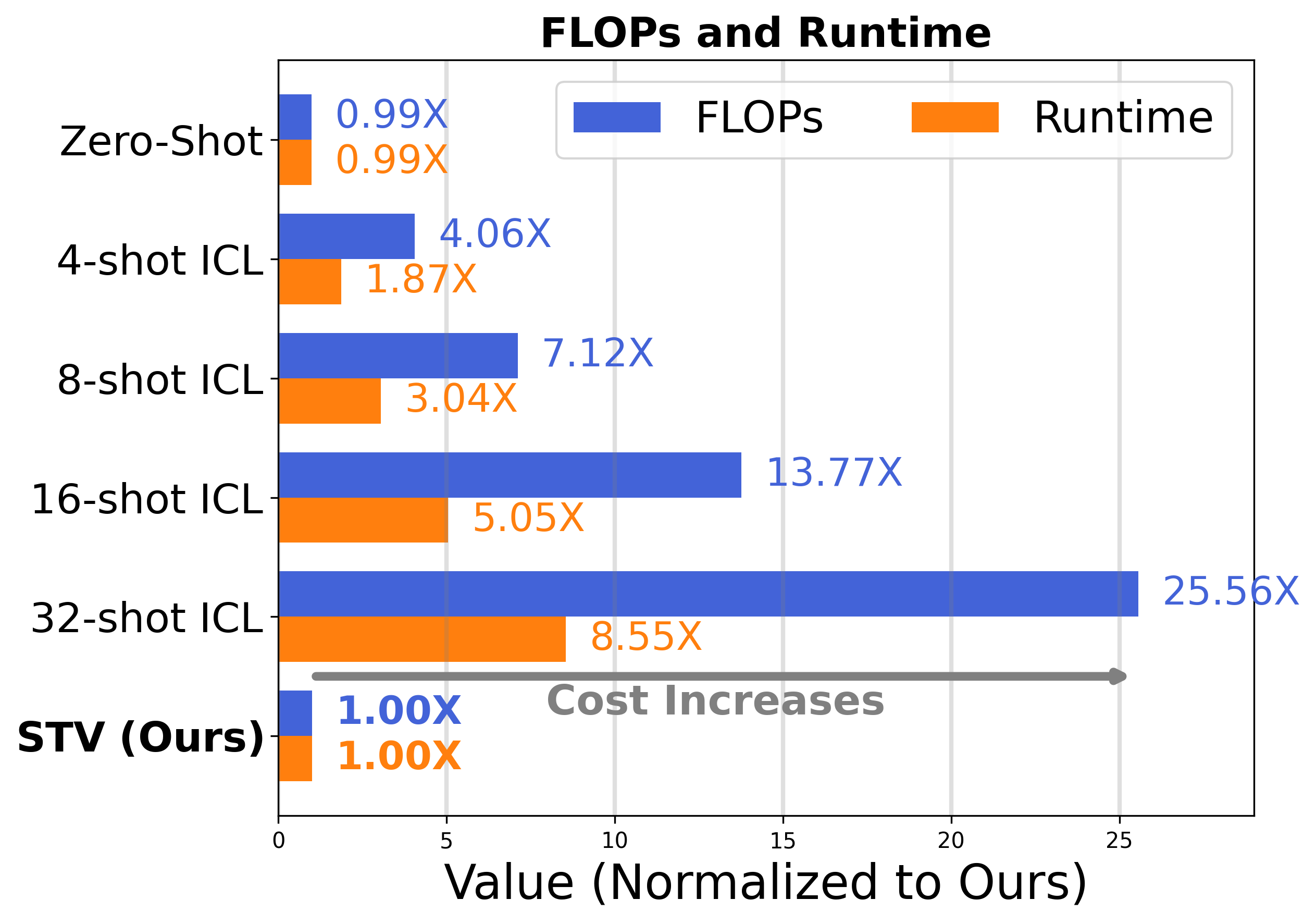} 
\caption{FLOPs and Runtime Comparison between STV and Few-Shot ICL.}
\label{fig:flop}
\end{figure}

\noindent\textbf{Efficiency and Effectiveness Compared with In-Context Learning.}
To validate our method, STV is compared with few-shot ICL baselines on VizWiz using Qwen-VL-7B. As shown in Figure~\ref{fig:flop}, STV achieves superior accuracy with negligible FLOPs and runtime overhead—remaining on par with zero-shot inference. In contrast, scaling ICL from 4-shot to 32-shot increases FLOPs up to 25$\times$ and runtime up to 8$\times$, highlighting the efficiency advantage of our insertion strategy under long-context constraints. Table~\ref{tab:icl_comparison} further benchmarks performance under varying shot counts. STV reaches 58.3\% accuracy, outperforming 32-shot ICL by +8.5\% while incurring no memory or token overhead. Moreover, 64-shot ICL fails due to out-of-memory (OOM), underscoring the limitation of naïve scaling. Overall, STV conveys task information more effectively than token-level prompting, with better efficiency and scalability.

\noindent\textbf{Effect of Exemplar Quality and Robustness.}
Table~\ref{aaai_tab:stv_resultsv2} shows that high-quality exemplars selected by Facility Location \cite{schreiber2020apricot} further improve STV (+3.6\%), demonstrating the benefit of exemplar quality in activation-space augmentation. Moreover, when inserting noisy cross-domain exemplars, STV exhibits smaller degradation ($-$0.7) than ICL ($-$1.0), indicating stronger robustness and stability.


\begin{table}[t]
\centering
\setlength{\tabcolsep}{10pt}
\renewcommand{\arraystretch}{1.2}
\begin{tabular}{l|cc}
\toprule
Model & VizWiz & OK-VQA \\
\midrule
Qwen-VL-7B         & 35.2  & 58.6 \\
+ \textbf{SFT}              & 62.0  {\small\ (+26.8)} & 25.1 {\small\ ($-$33.5)} \\
+ LoRA             & 44.3  {\small\ (+9.1)}  & 57.7 {\small\ ($-$0.9)} \\

STV (Ours)         & \textbf{58.3} {\small\ (+23.1)} & \textbf{61.9} {\small\ (+3.3)} \\
\bottomrule
\end{tabular}
\caption{Accuracy on VizWiz and OK-VQA using Qwen-VL with different adaptation strategies. The values in parentheses indicate the change compared to the base model. }
\label{tab:vizwiz_okvqa_comparison}
\end{table}

\begin{table}[t]
\centering
\setlength{\tabcolsep}{6pt}
\begin{tabular}{c|c|c|c|c}
\toprule
Method & Num. & OOM & Acc. (\%) & Change \\
\midrule
\multirow{6}{*}{\parbox{0.8cm}{\centering Standard\\ICL}}

  & 0   & $\times$ & 35.2  & --    \\
  & 4   & $\times$ & 42.0  & +6.8  \\
  & 8   & $\times$ & 44.3  & +9.1  \\
  & 16  & $\times$ & 46.9  & +11.7 \\
  & 32  & $\times$ & 49.8  & +14.6 \\
  & 64  & $\checkmark$ & --   & --    \\
\midrule
STV (Ours) & 400 & $\times$ & \textbf{58.3}  & \textbf{+23.1} \\
\bottomrule
\end{tabular}
\caption{Comparison of accuracy across different few-shot ICL settings and our proposed method (STV). Here “OOM” is the abbreviation for “out of memory”.}
\label{tab:icl_comparison}
\end{table}

\begin{table}[t]
\centering
\renewcommand{\arraystretch}{1.2}
\setlength{\tabcolsep}{10pt}

\begin{tabular}{l|c}
\toprule
Model & VizWiz Dataset \\
\midrule
\multicolumn{2}{c}{(a) STV with High-Quality Shots} \\
\midrule
Qwen-VL-7B              & 35.2 \\
+ STV                  & 58.3 (+23.1) \\
+ STV + F.L. Shots     & 61.9 (+26.7) \\
\midrule
\multicolumn{2}{c}{(b) Stability of ICL vs. STV} \\
\midrule
4-shot ICL             & 41.0 (-1.0) \\
STV                    & 57.6 (-0.7) \\
\bottomrule
\end{tabular}

\caption{(a) STV with high-quality shots and (b) Stability of ICL vs STV using Qwen-VL on VizWiz.}
\label{aaai_tab:stv_resultsv2}
\end{table}


\section{Related Work}
\textbf{Scaling In-Context Learning.} Early work \cite{brown2020language,bertsch2024context, li2023context,li2024long, chowdhery2023palm} show that more in-context examples improve language model performance, but gains were constrained by short context windows (e.g., 2048 in GPT-3) and typically fewer than 100 demonstrations. Recent studies scale ICL to 1,000+ demonstrations \cite{li2023context, agarwal2024many, chen2023extending, peng2023yarn}, achieving steady improvements across NLP tasks, though restricted to text-only settings without cross-model generalization \cite{ma2024gerea}.

\noindent\textbf{Multimodal In-Context Learning.} Research on multimodal ICL remains nascent. Closed-source LMMs (e.g., GPT-4V, Gemini) benefit from demonstrations in diverse vision-language tasks \cite{zhang2024out, han2023well}, and complex prompts improve relational reasoning \cite{zhao2023mmicl}. A recent benchmark \cite{jiang2024many} evaluates GPT-4o \cite{hurst2024gpt} and Gemini 1.5 Pro \cite{team2024gemini} with up to 2,000 multimodal shots, showing substantial gains across classification, VQA, and localization. However, these rely on full-sequence prompting with high inference cost, leaving efficiency questions open \cite{ma2025efficient, chen2025maple,jiang2025mimic, li2025implicit}.

\noindent\textbf{Task-Vector-based Methods.} Approaches based on task vectors  \cite{hendel2023context, liu2023context, todd2023function, huang2024multimodal, hojel2024finding, yang2025task} insert compact representations of demonstrations into activations, avoiding full-sequence prompts. Value Estimation methods \cite{hendel2023context, liu2023context, todd2023function} compute vectors (e.g., via PCA) and insert them at fixed locations, reducing cost but generalizing poorly to multimodal tasks. Location Selection methods \cite{huang2024multimodal} instead fix the vector and find insertion sites via policy networks, improving efficiency but suffering from task information loss and unstable selection.

\section{Conclusion}

This paper presents STV, a simple yet effective framework for enabling many-shot multimodal in-context learning without increasing input length or altering model weights. By locating context-sensitive attention heads and selecting task vectors via reinforcement learning, STV introduces task-specific knowledge through activation-level modulation. Extensive experiments across five benchmarks and two LMM families confirm consistent gains over previous ICL and task vector methods, achieving strong generalization and inference-time efficiency. These results underscore the value of sensitivity-aware activation control as a scalable adaptation strategy for large multimodal models.

\bibliography{aaai2026}
\setlength{\leftmargini}{20pt}
\makeatletter\def\@listi{\leftmargin\leftmargini \topsep .5em \parsep .5em \itemsep .5em}
\def\@listii{\leftmargin\leftmarginii \labelwidth\leftmarginii \advance\labelwidth-\labelsep \topsep .4em \parsep .4em \itemsep .4em}
\def\@listiii{\leftmargin\leftmarginiii \labelwidth\leftmarginiii \advance\labelwidth-\labelsep \topsep .4em \parsep .4em \itemsep .4em}\makeatother

\setcounter{secnumdepth}{0}
\renewcommand\thesubsection{\arabic{subsection}}
\renewcommand\labelenumi{\thesubsection.\arabic{enumi}}

\newcounter{checksubsection}
\newcounter{checkitem}[checksubsection]

\newcommand{\checksubsection}[1]{%
  \refstepcounter{checksubsection}%
  \paragraph{\arabic{checksubsection}. #1}%
  \setcounter{checkitem}{0}%
}

\newcommand{\checkitem}{%
  \refstepcounter{checkitem}%
  \item[\arabic{checksubsection}.\arabic{checkitem}.]%
}
\newcommand{\question}[2]{\normalcolor\checkitem #1 #2 \color{blue}}
\newcommand{\ifyespoints}[1]{\makebox[0pt][l]{\hspace{-15pt}\normalcolor #1}}

\newpage

\end{document}